\documentclass[10pt]{cai}

\graphicspath{{figs/}}

\usepackage{amsmath,amssymb,amsfonts}
\usepackage{algorithmic}
\usepackage{graphicx}
\usepackage{textcomp}
\usepackage{xcolor}
\usepackage[algo2e,linesnumbered,lined,boxed,vlined]{algorithm2e}
    
\def\b{\ensuremath\boldsymbol}

\begin{document}

\volumeheader{35}{0}
\begin{center}

  \title{Theoretical Connection between Locally Linear Embedding, Factor Analysis, and Probabilistic PCA}
  \maketitle

  \thispagestyle{empty}

    \begin{tabular}{cc}
    Benyamin Ghojogh\upstairs{\affilone,*}, Ali Ghodsi\upstairs{\affiltwo}, Fakhri Karray\upstairs{\affilone}, Mark Crowley\upstairs{\affilone}
   \\[0.25ex]
   {\small \upstairs{\affilone}Department of Electrical and Computer Engineering, University of Waterloo, ON, Canada} \\
   {\small \upstairs{\affiltwo}Department of Statistics and Actuarial Science, University of Waterloo, ON, Canada} 
  \end{tabular}
  
  \emails{
    \upstairs{*}Corresponding author: bghojogh@uwaterloo.ca 
    }
  \vspace*{0.2in}
\end{center}

\begin{abstract}
Locally Linear Embedding (LLE) is a nonlinear spectral dimensionality reduction and manifold learning method. It has two main steps which are linear reconstruction and linear embedding of points in the input space and embedding space, respectively. In this work, we look at the linear reconstruction step from a stochastic perspective where it is assumed that every data point is conditioned on its linear reconstruction weights as latent factors. The stochastic linear reconstruction of LLE is solved using expectation maximization. We show that there is a theoretical connection between three fundamental dimensionality reduction methods, i.e., LLE, factor analysis, and probabilistic Principal Component Analysis (PCA). The stochastic linear reconstruction of LLE is formulated similar to the factor analysis and probabilistic PCA. It is also explained why factor analysis and probabilistic PCA are linear and LLE is a nonlinear method. This work combines and makes a bridge between two broad approaches of dimensionality reduction, i.e., the spectral and probabilistic algorithms. 
\end{abstract}

\copyrightnotice

\section{Introduction}

Dimensionality reduction and manifold learning methods are widely useful for feature extraction, manifold unfolding, and data visualization.
The dimensionality reduction methods can be divided into three main categories, i.e., spectral methods, probabilistic methods, and neural network-based methods.
An example for spectral methods is Locally Linear Embedding (LLE) \cite{roweis2000nonlinear,saul2003think,ghojogh2020locally}. Examples for probabilistic methods are factor analysis \cite{fruchter1954introduction,child1990essentials} and probabilistic Principal Component Analysis (PCA) \cite{roweis1997algorithms,tipping1999probabilistic}. An example for neural network-based methods is variational autoencoder \cite{kingma2014auto} which formulates variational inference \cite{bishop2006pattern,blei2017variational,zhang2018advances,ghojogh2021factor} in an autoencoder framework. 
In this short paper, we show theoretical connections between the fundamental methods of LLE, factor analysis, and probabilistic PCA. Hence, we connect the spectral and probabilistic approaches of dimensionality reduction. 
We also explain why factor analysis and probabilistic PCA are linear and LLE is a nonlinear method.
In Section \ref{section_background}, we review the required background. Section \ref{section_stochastic_LLE} models the linear reconstruction of LLE stochastically. Finally, section \ref{section_conclusion} concludes the paper and discusses the connections of factor analysis, probabilistic PCA, and LLE algorithms.

\section{Technical Background and Preliminaries}\label{section_background}


\subsection{Marginal Multivariate Gaussian Distribution}

Consider two random variables $\b{x}_1 \in \mathbb{R}^{d_1}$ and $\b{x}_2 \in \mathbb{R}^{d_2}$ and let $\b{x}_3 := [\b{x}_1^\top, \b{x}_2^\top]^\top \in \mathbb{R}^{d_1 + d_2}$. Assume that $\b{x}_1$ and $\b{x}_2$ are jointly multivariate Gaussian, i.e., $\b{x}_3 \sim \mathcal{N}(\b{x}_3; \b{\mu}_3, \b{\Sigma}_3)$.  The mean and covariance can be decomposed as:
\begin{align}
&\b{\mu}_3 =  [\b{\mu}_1^\top, \b{\mu}_2^\top]^\top \in \mathbb{R}^{d_1 + d_2}, \quad \b{\Sigma}_3 = 
\begin{bmatrix}
\b{\Sigma}_{11} & \b{\Sigma}_{12} \\
\b{\Sigma}_{21} & \b{\Sigma}_{22}
\end{bmatrix}
\in \mathbb{R}^{(d_1 + d_2) \times (d_1 + d_2)}, 
\end{align}
where $\b{\mu}_1 \in \mathbb{R}^{d_1}$, $\b{\mu}_2 \in \mathbb{R}^{d_2}$, $\b{\Sigma}_{11} \in \mathbb{R}^{d_1 \times d_2}$, $\b{\Sigma}_{22} \in \mathbb{R}^{d_2 \times d_2}$, $\b{\Sigma}_{12} \in \mathbb{R}^{d_1 \times d_2}$, and $\b{\Sigma}_{21} = \b{\Sigma}_{12}^\top$. 

It can be shown that the marginal distributions for $\b{x}_1$ and $\b{x}_2$ are Gaussian distributions where $\mathbb{E}[\b{x}_1] = \b{\mu}_1$ and $\mathbb{E}[\b{x}_2] = \b{\mu}_2$. 
The covariance matrix of the joint distribution can be simplified as \cite{ghojogh2021factor}:
\begin{align}\label{equation_covariance_marginal}
&\b{\Sigma}_3 \! =\! \mathbb{E}\Bigg[
\begin{bmatrix}
(\b{x}_1 - \b{\mu}_1)(\b{x}_1 - \b{\mu}_1)^\top, (\b{x}_1 - \b{\mu}_1)(\b{x}_2 - \b{\mu}_2)^\top \\
(\b{x}_2 - \b{\mu}_2)(\b{x}_1 - \b{\mu}_1)^\top, (\b{x}_2 - \b{\mu}_2)(\b{x}_2 - \b{\mu}_2)^\top
\end{bmatrix}
\Bigg],
\end{align}
where $\mathbb{E}[.]$ is the expectation operator. 
According to the definition of the multivariate Gaussian distribution, the conditional distribution is also a Gaussian distribution, i.e., $\b{x}_2 | \b{x}_1 \sim \mathcal{N}(\b{x}_2; \b{\mu}_{x_2|x_1}, \b{\Sigma}_{x_2|x_1})$ where \cite{ghojogh2021factor}:
\begin{align}
&\mathbb{R}^{d_2} \ni \b{\mu}_{x_2|x_1} := \b{\mu}_2 + \b{\Sigma}_{21}\b{\Sigma}_{11}^{-1} (\b{x}_1 - \b{\mu}_1), \label{equation_mean_conditional_in_joint} \\
&\mathbb{R}^{d_2 \times d_2} \ni \b{\Sigma}_{x_2|x_1} := \b{\Sigma}_{22} - \b{\Sigma}_{21}\b{\Sigma}_{11}^{-1} \b{\Sigma}_{12}, \label{equation_covariance_conditional_in_joint}
\end{align}
and likewise we have for $\b{x}_1 | \b{x}_2 \sim \mathcal{N}(\b{x}_1; \b{\mu}_{x_1|x_2}, \b{\Sigma}_{x_1|x_2})$.
Also, note that the probability density function of $d$-dimensional Gaussian distribution is:
\begin{align}\label{equation_multivariate_Gaussian_PDF}
&\mathcal{N}(\b{x}; \b{\mu}, \b{\Sigma})\! =\! \frac{1}{\sqrt{(2\pi)^d |\b{\Sigma}|}} \exp\Big(\!\!- \frac{(\b{x} - \b{\mu})^\top \b{\Sigma}^{-1} (\b{x} - \b{\mu})}{2}\Big),
\end{align}
where $|.|$ denotes the determinant of matrix. 


\begin{figure}[!t]
\centering
\includegraphics[width=0.35in]{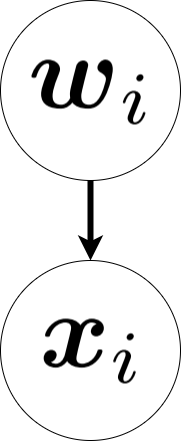}
\caption{The probabilistic graphical model for stochastic linear reconstruction in LLE.}
\label{figure_PGM}
\end{figure}

\subsection{Variational Inference}

Assume data $\b{x}_i$ is conditioned on a latent factor $\b{w}_i$, as shown in Fig. \ref{figure_PGM}. Let the parameters of model be denoted by $\b{\theta}$.
In variational inference, the Evidence Lower Bound (ELBO) is a lower bound on the log likelihood of data and is defined as minus Kullback-Leibler (KL) divergence between a distribution $q(.)$ on $\b{w}_i$ and the joint distribution of $\b{x}_i$ and $\b{w}_i$, i.e., $\mathcal{L}(q, \b{\theta}) := - \text{KL}\big(q(\b{w}_i)\, \|\, \mathbb{P}(\b{x}_i, \b{w}_i\, |\, \b{\theta})\big)$ \cite{ghojogh2021factor,bishop2006pattern,kingma2014auto}.
Maximizing this lower bound results in maximization of likelihood of data \cite{blei2017variational,zhang2018advances}. 
Variational inference uses EM for Maximum Likelihood Estimation (MLE) \cite{ghojogh2021factor}:
\begin{align}
& q^{(t)}(\b{w}_i) \gets \mathbb{P}(\b{w}_i\, |\, \b{x}_i, \b{\theta}^{(t-1)}), \label{equation_E_step_variationalInference} \\
& \b{\theta}^{(t)} \gets \arg \max_{\b{\theta}}~ \mathbb{E}_{\sim q^{(t)}(\b{w}_i)} \big[\log \mathbb{P}(\b{x}_i, \b{w}_i\, |\, \b{\theta})\big], \label{equation_M_step_variationalInference}
\end{align}
where $\mathbb{E}[.]$ denotes expectation.
Here, we use the EM approach of variational inference for stochastic linear reconstruction in LLE. 

\subsection{Factor Analysis}

Factor analysis \cite{fruchter1954introduction,child1990essentials} assumes that every data point $\b{x}_i$ is generated from a latent factor $\b{w}_i$. Its probabilistic graphical model is similar to Fig. \ref{figure_PGM} but with a small difference \cite{ghahramani1996algorithm}. 
It assumes $\b{x}_i$ is obtained by linear projection of $\b{w}_i$ onto the space of data by projection matrix $\b{\Lambda}$, then applying some linear translation, and finally adding a Gaussian noise $\b{\epsilon}$ with covariance matrix $\b{\Psi}$. This addition of noise is the main difference from the model depicted in Fig. \ref{figure_PGM}. If $\b{\mu}$ denotes the mean of data, factor analysis considers:
\begin{align}
&\b{x}_i := \b{\Lambda} \b{w}_i + \b{\mu} + \b{\epsilon}, \label{equation_factor_analysis} \\
&\mathbb{P}(\b{x}_i\, |\, \b{w}_i, \b{\Lambda}, \b{\mu}, \b{\Psi}) = \mathcal{N}(\b{x}_i; \b{\Lambda} \b{w}_i + \b{\mu}, \b{\Psi}), \label{equation_factor_analysis_likelihood_x_given_z}
\end{align}
where $\mathbb{P}(\b{w}_i) = \mathcal{N}(\b{w}_i; \b{0}, \b{I})$ and $\mathbb{P}(\b{\epsilon}) = \mathcal{N}(\b{\epsilon}; \b{0}, \b{\Psi})$. Factor analysis uses EM algorithm for finding optimum $\b{\Lambda}$ and $\b{\Psi}$ (see \cite{ghojogh2021factor} for details of EM in factor analysis).
We will show that the stochastic linear reconstruction in LLE, for modeling the relation between a data point and its latent reconstruction weights, is similar to Eq. (\ref{equation_factor_analysis}).

\subsection{Probabilistic PCA}

Probabilistic PCA \cite{roweis1997algorithms,tipping1999probabilistic} is a special case of factor analysis where the variance of noise is equal in all dimensions of data space with covariance between dimensions, i.e.:
\begin{align}\label{equation_PPCA_Psi}
\b{\Psi} = \sigma^2 \b{I}.
\end{align}
In other words, probabilistic PCA considers an isotropic noise model.
Similar to factor analysis, it can be solved iteratively using EM \cite{roweis1997algorithms}. However, one can also find a closed-form solution to its EM approach \cite{tipping1999probabilistic}.
Hence, by restricting the noise covariance to be isotropic, its solution becomes simpler and closed-form. See \cite{ghojogh2021factor} for details of derivations and solutions for probabilistic PCA. 


\section{Stochastic Modeling of Linear Reconstruction in LLE}\label{section_stochastic_LLE}




\subsection{Notations and Joint and Conditional Distributions}

LLE \cite{roweis2000nonlinear,saul2003think} has two main steps which are linear reconstruction and linear embedding \cite{ghojogh2020locally}. 
As Fig. \ref{figure_PGM} depicts, the linear reconstruction step of LLE can be seen stochastically where every point $\b{x}_i$ is conditioned on and generated by its reconstruction weights $\b{w}_i$ as a latent factor. 
Therefore, $\b{x}_i$ can be written as a stochastic function of $\b{w}_i$ where we assume $\b{w}_i$ has a multivariate Gaussian distribution:
\begin{align}
&\b{x}_i = \b{X}_i \b{w}_i + \b{\mu}, \label{equation_x_X_w} \\
&\mathbb{P}(\b{w}_i) = \mathcal{N}(\b{w}_i; \b{0}, \b{\Omega}_i) \implies \mathbb{E}[\b{w}_i] = \b{0}, ~~\mathbb{E}[\b{w}_i \b{w}_i^\top] = \b{\Omega}_i, \label{equation_prior_w} 
\end{align}
where $\b{\Omega}_i \in \mathbb{R}^{k \times k}$ is covariance of $\b{w}_i$ and $\b{\mu} \in \mathbb{R}^d$ is the mean of data because $(1/n) \sum_{i=1}^n \b{x}_i = \mathbb{E}[\b{x}_i] = \b{X}_i \mathbb{E}[\b{w}_i] + \b{\mu} \overset{(\ref{equation_prior_w})}{=} \b{0} + \b{\mu} = \b{\mu}$.
According to Eq. (\ref{equation_covariance_marginal}), for the joint distribution of $[\b{x}_i^\top, \b{w}_i^\top]^\top \in \mathbb{R}^{d + k}$,  we have:
\begin{alignat*}{2}
&\b{\Sigma}_{11} &&= \mathbb{E}[(\b{x}_i - \b{\mu}) (\b{x}_i - \b{\mu})^\top] \overset{(\ref{equation_x_X_w})}{=} \mathbb{E}[(\b{X}_i \b{w}_i) (\b{X}_i \b{w}_i)^\top] = \b{X}_i \mathbb{E}[\b{w}_i \b{w}_i^\top] \b{X}_i^\top \overset{(\ref{equation_prior_w})}{=} \b{X}_i \b{\Omega}_i \b{X}_i^\top, \\
&\b{\Sigma}_{12} &&= \mathbb{E}[(\b{x}_i - \b{\mu}) (\b{w}_i - \b{0})^\top] \overset{(\ref{equation_x_X_w})}{=} \b{X}_i \mathbb{E}[\b{w}_i \b{w}_i^\top] \overset{(\ref{equation_prior_w})}{=} \b{X}_i \b{\Omega}_i, \\
&\b{\Sigma}_{22} &&= \mathbb{E}[(\b{w}_i - \b{0}) (\b{w}_i - \b{0})^\top] = \mathbb{E}[\b{w}_i \b{w}_i^\top] \overset{(\ref{equation_prior_w})}{=} \b{\Omega}_i.
\end{alignat*}
Hence:
\begin{align}\label{equation_joint_distribution_x_w}
\begin{bmatrix}
\b{x}_{i} \\
\b{w}_{i}
\end{bmatrix}
\sim
\mathcal{N}\Bigg(
\begin{bmatrix}
\b{x}_i \\
\b{w}_i 
\end{bmatrix}
;
\begin{bmatrix}
\b{\mu} \\
\b{0} 
\end{bmatrix}
,
\begin{bmatrix}
\b{X}_i \b{\Omega}_i \b{X}_i^\top & \b{X}_i \b{\Omega}_i \\
\b{\Omega}_i^\top \b{X}_i^\top & \b{\Omega}_i
\end{bmatrix}
\Bigg).
\end{align}
We have:
\begin{align}\label{equation_x_given_w}
&\mathbb{P}(\b{x}_i\, |\, \b{w}_i, \b{\Omega}_i) \overset{(a)}{=} \mathcal{N}(\b{x}_i; \b{X}_i \b{w}_i + \b{\mu}, \b{X}_i \b{\Omega}_i \b{X}_i^\top),
\end{align}
where $(a)$ is because of Eqs. (\ref{equation_x_X_w}) and (\ref{equation_joint_distribution_x_w}).
We can use EM for MLE in stochastic linear reconstruction of LLE. In the following, the steps of EM are explained.  

\subsection{E-Step in Expectation Maximization}

As we will see later in the M-step of EM, we will have two expectation terms which need to be computed in the E-step. These expectations, which are over the $q(\b{w}_i) := \mathbb{P}(\b{w}_i\, |\, \b{x}_i)$ distribution, are $\mathbb{E}_{\sim q^{(t)}(\b{w}_i)}[\b{w}_i] \in \mathbb{R}^k$ and $\mathbb{E}_{\sim q^{(t)}(\b{w}_i)}[\b{w}_i \b{w}_i^\top] \in \mathbb{R}^{k \times k}$ where $t$ denotes the iteration index in EM iterations.
According to Eqs. (\ref{equation_mean_conditional_in_joint}), (\ref{equation_covariance_conditional_in_joint}), and (\ref{equation_joint_distribution_x_w}), we have:
\begin{align}
&\mathbb{E}_{\sim q^{(t)}(\b{w}_i)}[\b{w}_i] = \b{\mu}_{w|x} = \b{\Omega}_i^\top \b{X}_i^\top (\b{X}_i \b{\Omega}_i \b{X}_i^\top)^{\dagger} (\b{x}_i - \b{\mu}), \label{E_step_expectation_w} \\
&\mathbb{E}_{\sim q^{(t)}(\b{w}_i)}[\b{w}_i \b{w}_i^\top] = \b{\Sigma}_{w|x} = \b{\Omega}_i - \b{\Omega}_i^\top \b{X}_i^\top (\b{X}_i \b{\Omega}_i \b{X}_i^\top)^{\dagger} \b{X}_i \b{\Omega}_i, \label{E_step_expectation_w_w} 
\end{align}
where $^\dagger$ denotes either inverse or pseudo-inverse of matrix. 

\subsection{M-Step in Expectation Maximization}

In M-step of EM, we maximize the joint likelihood of data and weights over all $n$ data points where the optimization variable is the covariance of prior distribution of weights:
\begin{align}
& \max_{\{\b{\Omega}_i\}_{i=1}^n}~ \sum_{i=1}^n \mathbb{E}_{\sim q^{(t)}(\b{w}_i)} \big[\log \mathbb{P}(\b{x}_i, \b{w}_i\, |\, \b{\Omega}_i)\big] \nonumber \\
&\overset{(a)}{=} \max_{\{\b{\Omega}_i\}_{i=1}^n}~ \sum_{i=1}^n \Big( \mathbb{E}_{\sim q^{(t)}(\b{w}_i)} \big[\log \mathbb{P}(\b{x}_i\, |\, \b{w}_i, \b{\Omega}_i)\big] + \mathbb{E}_{\sim q^{(t)}(\b{w}_i)} \big[\log \mathbb{P}(\b{w}_i)\big] \Big) \nonumber\\
& \overset{(b)}{=} \max_{\{\b{\Omega}_i\}_{i=1}^n} \sum_{i=1}^n \Big( \mathbb{E}_{\sim q^{(t)}(\b{w}_i)} \big[\log \mathcal{N}(\b{X}_i \b{w}_i + \b{\mu}, \b{X}_i \b{\Omega}_i \b{X}_i^\top) \big]   + \mathbb{E}_{\sim q^{(t)}(\b{w}_i)} \big[\log \mathcal{N}(\b{0}, \b{\Omega}_i) )\big] \Big) \nonumber
\end{align}
\begin{align}
&\overset{(\ref{equation_multivariate_Gaussian_PDF})}{=} \max_{\{\b{\Omega}_i\}_{i=1}^n}~ \Big( \underbrace{-\frac{d\,n}{2} \log(2\pi)}_\text{constant} - \frac{n}{2} \log |\b{X}_i \b{\Omega}_i \b{X}_i^\top| \nonumber \\
&- \frac{1}{2} \sum_{i=1}^n \mathbb{E}_{\sim q^{(t)}(\b{w}_i)} \big[ (\b{x}_i - \b{X}_i\b{w}_i - \b{\mu})^\top (\b{X}_i \b{\Omega}_i \b{X}_i^\top)^{-1}  (\b{x}_i - \b{X}_i\b{w}_i - \b{\mu}) \big] \nonumber\\
&-\underbrace{\frac{k\,n}{2} \log(2\pi)}_\text{constant}  - \frac{n}{2} \log |\b{\Omega}_i| - \frac{1}{2} \sum_{i=1}^n \mathbb{E}_{\sim q^{(t)}(\b{w}_i)} \big[ \b{w}_i^\top \b{\Omega}_i^{-1} \b{w}_i \big] \Big) \nonumber\\
&\overset{(c)}{=} \max_{\{\b{\Omega}_i\}_{i=1}^n} \Big(\!\! - \frac{n}{2} \log |\b{X}_i \b{\Omega}_i \b{X}_i^\top| - \frac{n}{2} \textbf{tr}\big((\b{X}_i \b{\Omega}_i \b{X}_i^\top)^{-1} \b{S}_1\big)  - \frac{n}{2} \log |\b{\Omega}_i| - \frac{n}{2} \textbf{tr}\big( \b{\Omega}_i^{-1} \b{S}_2 \big) \Big), \label{equation_joint_likelihood}
\end{align}
where $(a)$ is because of the chain rule $\mathbb{P}(\b{x}_i, \b{w}_i\, |\, \b{\Omega}_i) = \mathbb{P}(\b{x}_i\, |\, \b{w}_i, \b{\Omega}_i)\, \mathbb{P}(\b{w}_i)$, and $(b)$ is because of Eqs. (\ref{equation_prior_w}) and (\ref{equation_x_given_w}), and $(c)$ is because we define the scatters $\b{S}_1$ and $\b{S}_2$ as:
\begin{align}
&\mathbb{R}^{d \times d} \ni \b{S}_1 := \frac{1}{n} \sum_{i=1}^n \mathbb{E}_{\sim q^{(t)}(\b{w}_i)} \big[ (\b{x}_i - \b{X}_i\b{w}_i - \b{\mu})  (\b{x}_i - \b{X}_i\b{w}_i - \b{\mu})^\top \big] \quad\quad\quad\quad\quad\quad\quad\quad \nonumber
\end{align}
\begin{align}
&= \frac{1}{n} \sum_{i=1}^n \Big( (\b{x}_i - \b{\mu}) (\b{x}_i - \b{\mu})^\top  - 2 \b{X}_i \mathbb{E}_{\sim q^{(t)}(\b{w}_i)} \big[ \b{w}_i \big] (\b{x}_i - \b{\mu})^\top  + \b{X}_i \mathbb{E}_{\sim q^{(t)}(\b{w}_i)} \big[ \b{w}_i \b{w}_i^\top \big] \b{X}_i^\top \Big), \label{equation_S1}
\end{align}
\begin{align}
&\mathbb{R}^{k \times k} \ni \b{S}_2 := \frac{1}{n} \sum_{i=1}^n \mathbb{E}_{\sim q^{(t)}(\b{w}_i)} \big[ \b{w}_i \b{w}_i^\top \big], \label{equation_S2}
\end{align}
where the expectation terms are found by Eqs. (\ref{E_step_expectation_w}) and (\ref{E_step_expectation_w_w}). 
The gradient of the joint likelihood is:
\begin{equation}\label{equation_gradient_joint_likelihood}
\begin{aligned}
\mathbb{R}^{k \times k} \ni \frac{\partial \text{ Eq. } (\ref{equation_joint_likelihood})}{\partial \b{\Omega}_i^{-1}} = &\frac{n}{2} \Big[ \textbf{vec}^{-1}_{k \times k} \big[ \b{T}_i\, \textbf{vec}_{d^2 \times 1}(\b{X}_i \b{\Omega}_i \b{X}_i^\top) \big] \\
&- \textbf{vec}^{-1}_{k \times k} \big[ \b{T}_i\, \textbf{vec}_{d^2 \times 1}(\b{S}_1) \big] + \b{\Omega}_i - \b{S}_2 \Big],
\end{aligned}
\end{equation}
where we use the Magnus-Neudecker convention in which matrices are vectorized, $\textbf{vec}(.)$ vectorizes the matrix, $\textbf{vec}^{-1}_{k \times k}(.)$ is de-vectorization to $k \times k$ matrix, $\otimes$ denotes the Kronecker product, and $\mathbb{R}^{k^2 \times d^2} \ni \b{T}_i := \b{X}_i^\top \otimes \b{X}_i^\top$.
In the M-step, one can update the variables $\{\b{\Omega}_i\}_{i=1}^n$ using gradient descent with the gradient in Eq. (\ref{equation_gradient_joint_likelihood}). However, we can relax the covariance matrix and simplify the algorithm. 

\subsection{Relaxation of Covariance}

Inspired by relaxation of factor analysis for probabilistic PCA (see Eq. (\ref{equation_PPCA_Psi})), we can relax the covariance matrix to be spherical, i.e., diagonal and the variance of weights to be equal in all $k$ dimensions: 
\begin{align}\label{equation_diagonal_Omega_covariance}
\b{\Omega}_i = \sigma_i \b{I} \in \mathbb{R}^{k \times k}.
\end{align}
Substituting this covariance into Eq. (\ref{equation_joint_likelihood}) and noticing the properties of determinant and trace gives:
\begin{align}
&\max_{\{\sigma_i\}_{i=1}^n} \Big(\!\! - \frac{n}{2} \log (\sigma_i^d |\b{X}_i \b{X}_i^\top|) - \frac{n}{2} \sigma_i^{-1} \textbf{tr}\big((\b{X}_i \b{X}_i^\top)^{-1} \b{S}_1\big)  - \frac{n}{2} \log \sigma_i^k - \frac{n}{2} \sigma_i^{-1} \textbf{tr}\big( \b{S}_2 \big) \Big) \nonumber \\
&\overset{(a)}{=} \max_{\{\sigma_i\}_{i=1}^n} \Big(\!\! - \frac{n}{2} (d + k) \log (\sigma_i)  - \frac{n}{2} \Big[ \textbf{tr}\big((\b{X}_i \b{X}_i^\top)^{-1} \b{S}_1\big) - \textbf{tr}\big( \b{S}_2 \big) \Big] \sigma_i^{-1} \Big), \label{equation_joint_likelihood_relaxed}
\end{align}
where $(a)$ is because $\log (\sigma_i^d |\b{X}_i \b{X}_i^\top|) = d\log (\sigma_i) + \log (|\b{X}_i \b{X}_i^\top|)$ whose second term is a constant w.r.t. $\sigma_i$. 
Setting the gradient of the joint likelihood to zero gives:
\begin{align}
&\mathbb{R} \ni \frac{\partial \text{ Eq. } (\ref{equation_joint_likelihood_relaxed})}{\partial \sigma_i^{-1}} = \frac{n}{2} \Big[ (d + k) \sigma_i  - \Big( \textbf{tr}\big((\b{X}_i \b{X}_i^\top)^{-1} \b{S}_1\big) + \textbf{tr}(\b{S}_2) \Big) \Big] \overset{\text{set}}{=} 0 \nonumber \\
&\implies \sigma_i = (d + k)^{-1} \Big( \textbf{tr}\big((\b{X}_i \b{X}_i^\top)^{\dagger} \b{S}_1\big) + \textbf{tr}(\b{S}_2) \Big), \label{equation_variance_M_step}
\end{align}
where $^\dagger$ denotes either inverse or pseudo-inverse of matrix. 
As we also have in probabilistic PCA, the relaxation of covariance matrix results in the closed-form solution of Eq. (\ref{equation_variance_M_step}). 
Without this relaxation, the solution of M-step in EM algorithm of LLE is solved iteratively by the gradient in Eq. (\ref{equation_gradient_joint_likelihood}). 

The EM algorithm for stochastic linear reconstruction in LLE is summarized in Algorithm \ref{algorithm_linear_reconstruction_EM}.
We can sample $\{\b{w}_i\}_{i=1}^n$ with the following prior and conditional distributions:
\begin{align}
& \b{w}_i \sim \mathcal{N}(\b{w}_i; \b{0}, \sigma_i \b{I}), \\
& \b{w}_i\, |\, \b{x}_i \sim \mathcal{N}(\b{w}_i; \b{\mu}_{w | x}, \b{\Sigma}_{w | x}), \label{equation_distribution_w_given_x}
\end{align}
where $\b{\mu}_{w | x}$ and $\b{\Sigma}_{w | x}$ are defined in Eqs. (\ref{E_step_expectation_w}) and (\ref{E_step_expectation_w_w}), respectively. 


\SetAlCapSkip{0.5em}
\IncMargin{0.8em}
\begin{algorithm2e}[!t]
\footnotesize
\DontPrintSemicolon
    \textbf{Input: } $\{\b{x}_i\}_{i=1}^n$, $k$NN graph or $\{\b{X}_i\}_{i=1}^n$ \;
    \textbf{Initialize: } $\{\b{\Omega}_i\}_{i=1}^n = \b{I}$\;
    \While{not converged}{
        // E-step:\;
        \For{every data point $i$ from $1$ to $n$}{
            Calculate expectations by Eqs. (\ref{E_step_expectation_w}) and (\ref{E_step_expectation_w_w})\;
        }
        // Sampling:\;
        Sample weights $\{\b{w}_i\}_{i=1}^n$ using Eq. (\ref{equation_distribution_w_given_x})\;
        // M-step:\;
        Calculate $\b{S}_1$ and $\b{S}_2$ using Eqs. (\ref{equation_S1}) and (\ref{equation_S2})\;
        \For{every data point $i$ from $1$ to $n$}{
            Calculate $\sigma_i$ by Eq. (\ref{equation_variance_M_step})\;
            Calculate $\b{\Omega}_i$ using Eq. (\ref{equation_diagonal_Omega_covariance})\;
        }
    }
    \textbf{Return} weights $\{\b{w}_i\}_{i=1}^n$\;
\caption{Stochastic Linear Reconstruction in LLE with EM}\label{algorithm_linear_reconstruction_EM}
\end{algorithm2e}
\DecMargin{0.8em}

\section{Conclusion and Discussion on the Connection of Methods}\label{section_conclusion}

Comparing Eqs. (\ref{equation_factor_analysis}) and (\ref{equation_x_X_w}) shows that data point $\b{x}_i$ is conditioned on some latent variable $\b{w}_i$, in all methods of factor analysis \cite{fruchter1954introduction,child1990essentials}, probabilistic PCA \cite{roweis1997algorithms,tipping1999probabilistic}, and LLE \cite{roweis2000nonlinear,saul2003think}. In these methods, the latent variable $\b{w}_i$ is related to data $\b{x}_i$ using a transformation matrix. 
In factor analysis and probabilistic PCA (see Eq. (\ref{equation_factor_analysis})), this transformation matrix is a global matrix $\b{\Lambda}$ so it is data-independent in the sense that it is the same matrix for all data points. However, the transformation matrix of LLE is $\b{X}_i$ (see Eq. (\ref{equation_x_X_w})) which is local and data-dependent in the sense that it is different for every data point. This explains why factor analysis and probabilistic PCA are linear methods and LLE is a nonlinear algorithm. 
Moreover, in this framework, if the covariance matrix is relaxed to be spherical (see Eqs. (\ref{equation_PPCA_Psi}) and (\ref{equation_diagonal_Omega_covariance})), the solution becomes closed-form. This relaxation happens in probabilistic PCA and can also be used in LLE to have a closed-form solution (see Eq. (\ref{equation_variance_M_step})). 
Finally, the introduced relation of the three algorithms opens the door for more investigation in relation of spectral and probabilistic approaches of machine learning.


\printbibliography[heading=subbibintoc]

@article{ghojogh2021factor,
  title={Factor Analysis, Probabilistic Principal Component Analysis, Variational Inference, and Variational Autoencoder: Tutorial and Survey},
  author={Ghojogh, Benyamin and Ghodsi, Ali and Karray, Fakhri and Crowley, Mark},
  journal={arXiv preprint arXiv:2101.00734},
  year={2021}
}

@article{roweis2000nonlinear,
  title={Nonlinear dimensionality reduction by locally linear embedding},
  author={Roweis, Sam T and Saul, Lawrence K},
  journal={Science},
  volume={290},
  number={5500},
  pages={2323--2326},
  year={2000},
  publisher={American Association for the Advancement of Science}
}

@article{saul2003think,
  title={Think globally, fit locally: unsupervised learning of low dimensional manifolds},
  author={Saul, Lawrence K and Roweis, Sam T},
  journal={Journal of machine learning research},
  volume={4},
  number={Jun},
  pages={119--155},
  year={2003}
}

@article{ghojogh2020locally,
  title={Locally Linear Embedding and its Variants: Tutorial and Survey},
  author={Ghojogh, Benyamin and Ghodsi, Ali and Karray, Fakhri and Crowley, Mark},
  journal={arXiv preprint arXiv:2011.10925},
  year={2020}
}

@book{bishop2006pattern,
  title={Pattern recognition and machine learning},
  author={Bishop, Christopher M},
  year={2006},
  publisher={Springer}
}

@techreport{ghahramani1996algorithm,
  title={The {EM} algorithm for mixtures of factor analyzers},
  author={Ghahramani, Zoubin and Hinton, Geoffrey E},
  year={1996},
  institution={Technical Report CRG-TR-96-1, University of Toronto}
}

@book{child1990essentials,
  title={The essentials of factor analysis},
  author={Child, Dennis},
  year={1990},
  publisher={Cassell Educational}
}

@book{fruchter1954introduction,
  title={Introduction to factor analysis},
  author={Fruchter, Benjamin},
  year={1954},
  publisher={Van Nostrand}
}

@article{roweis1997algorithms,
  title={{EM} algorithms for {PCA} and {SPCA}},
  author={Roweis, Sam},
  journal={Advances in neural information processing systems},
  volume={10},
  pages={626--632},
  year={1997}
}

@article{tipping1999probabilistic,
  title={Probabilistic principal component analysis},
  author={Tipping, Michael E and Bishop, Christopher M},
  journal={Journal of the Royal Statistical Society: Series B (Statistical Methodology)},
  volume={61},
  number={3},
  pages={611--622},
  year={1999},
  publisher={Wiley Online Library}
}

@inproceedings{kingma2014auto,
  title={Auto-encoding variational {Bayes}},
  author={Kingma, Diederik P and Welling, Max},
  booktitle={International Conference on Learning Representations},
  year={2014}
}

@article{blei2017variational,
  title={Variational inference: A review for statisticians},
  author={Blei, David M and Kucukelbir, Alp and McAuliffe, Jon D},
  journal={Journal of the American statistical Association},
  volume={112},
  number={518},
  pages={859--877},
  year={2017},
  publisher={Taylor \& Francis}
}

@article{zhang2018advances,
  title={Advances in variational inference},
  author={Zhang, Cheng and B{\"u}tepage, Judith and Kjellstr{\"o}m, Hedvig and Mandt, Stephan},
  journal={IEEE transactions on pattern analysis and machine intelligence},
  volume={41},
  number={8},
  pages={2008--2026},
  year={2018},
  publisher={IEEE}
}

\end{document}